 \documentclass[sigconf,10pt]{acmart}

\usepackage{subcaption}
\usepackage{enumitem}
\usepackage{amsfonts,amsmath,amsthm}
\usepackage{xurl}

\usepackage{algorithm} \usepackage{algpseudocode}

\usepackage[normalem]{ulem}

\AtBeginDocument{%
  }

\renewcommand\footnotetextcopyrightpermission[1]{} 

\newcommand{\xref}[1]{\S\ref{#1}}

\newcommand{\squishlist}{\begin{itemize}[itemsep=1pt,parsep=2pt,topsep=3pt,partopsep=0pt,leftmargin=0em, itemindent=1em,labelwidth=1em,labelsep=0.5em]}
\newcommand{\squishend}{\end{itemize}}

\acmConference[Mobile AI Factory]{}{October}{2026}

\setcopyright{none}                 
\renewcommand\footnotetextcopyrightpermission[1]{} 

\begin{document}

\title{Mobile AI Factory: Small Language Model Creation Agent for Mobile Devices}

\title{\huge{Mobile AI Factory: From Task Descriptions to On-Device Language Models}}

\title{\huge{From Prompt to Phone: A Mobile AI  Factory for On-Device Language Models}}

\title{\huge{From Prompt to Phone:  Specialized Language Model Factory}}

\title{\huge{Specialized Language Model Factory: From Prompt to Phone}}

\title{\huge{Specialized Language Model Factory: From Prompt to Phone}}

\title{\huge{Specialized Language Model Factory for Mobile Devices}}

\title{No Model Fits All Phones: An Agentic Factory for On-Device Language Models}

\title{An Agentic Factory for On-Device Language Models}

\title{Mobile AI Factory: Agentic Production of  On-Device Specialized Language Models}

\title{Mobile AI Factory: Agentic Creation of Specialized On-Device Language Models}


\title{Mobile AI Factory: Agents That Build Specialized Language Models for Your Phone}

\title{{Mobile AI Factory: Agents That Build Specialized Small Language Models}}

\title{Mobile AI Factory: Building Specialized   Language Models for Your Mobile Device}


\newif\ifshowrevisions \showrevisionstrue
\ifshowrevisions
  \DeclareRobustCommand{\revdel}[1]{\sout{#1}}
  \DeclareRobustCommand{\revadd}[1]{{\color{red}#1}}
  \DeclareRobustCommand{\revchg}[2]{\sout{#1} {\color{red}#2}}
\else
  \DeclareRobustCommand{\revdel}[1]{}
  \DeclareRobustCommand{\revadd}[1]{#1}
  \DeclareRobustCommand{\revchg}[2]{#2}
\fi

\author{Evan Li$^*$, Krishna Deshpande$^*$, Wen Cheng$^*$ and Shyamnath Gollakota}
\affiliation{
\institution{Paul G. Allen School of Computer Science and Engineering, University of  Washington}
\institution{\{evanly, kdeshpan, wencheng, gshyam\} @cs.washington.edu}
\institution{$^*$co-primary authors}
\country{}
}







\begin{abstract}
Small language models (SLMs) running on mobile devices can keep user data
private and work without connectivity. But they perform well only when
fine-tuned for the target task and sized for the target device, a process that
today requires repeated expert tuning. We present {\it Mobile AI Factory}, an
agentic system that builds a fine-tuned, quantized SLM for a given task and
phone. Given a task specification and a target phone, an orchestrator LLM
chooses whether to tune hyperparameters, mine real data, or generate synthetic
examples, while fixed rules roll back changes that hurt accuracy. The factory
searches from small to large models across base models and quantizations,
adapts its accuracy goal as it searches, and uses real measurements of the
target phone across inference engines and backends. We evaluate Mobile AI
Factory on ten tasks and four Android phones. On an unseen testset, its on-device models outperform two five-shot  frontier
models on five tasks, by up to 0.34, and match them on
three more. Its search comes within 0.03 of exhaustively training all 27 model
variants on an entity extraction task, while using 19$\times$ less GPU time.
Phone measurements also matter: the best configuration changes across phones. Without phone measurements,  Claude Opus 5.5, given the full hardware and model specifications, misses the app's requirements in 25\% of easy and
moderate cases and in 60\% of hard ones.  Code is available at:  \url{https://github.com/evanly-gh/SLM_Factory}.


\end{abstract}

\maketitle

\providecommand{\new}[1]{\textcolor{red}{#1}}
 
\section{Introduction}
\label{sec:intro}
 
Small language models that run directly on mobile devices could enable numerous  applications. For example, a keyboard  app could rewrite a hastily typed message into a more professional response, an email client could summarize messages and draft replies, and a voice assistant could help users with motor impairments fill out forms and navigate applications. Running these models on the device, rather than in the cloud, has two key advantages. First, sensitive data, like a patient's medication schedule or a customer's bank details, can remain on the device. Second, applications can operate in scenarios  where connectivity is limited or intermittent, such as translating a conversation on a hiking trail or providing navigation when a vehicle loses network access.
 
\begin{figure}[t!]
  \centering
\vskip -0.15in
  \includegraphics[width=0.85\columnwidth]{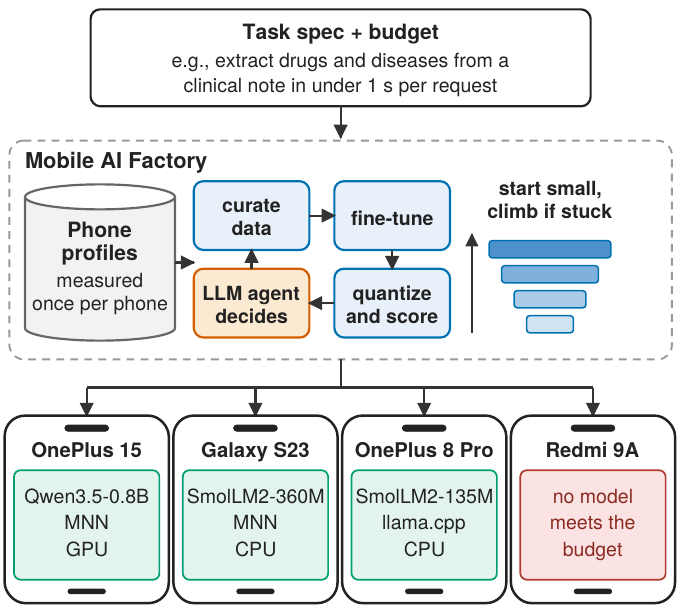}
  \vskip -0.15in
  \caption{Mobile AI Factory. \textnormal{It takes a task specification and a budget, and builds a specialized model for a phone.}}
  \vskip -0.15in
  \label{fig:concept}
\end{figure}
 
Smartphones now include capable CPUs and GPUs, and mobile inference engines
such as llama.cpp and MNN can run quantized SLMs at practical
speeds~\cite{llamacpp}. The harder problem is building the models
themselves. Each application needs a model that is trained for its task and
small enough to run on the target phone. Even with training data available, developers still need to fine-tune
a base model, iterate on accuracy, decide on the model size and quantization
level, and test the resulting model on the target phone.
 
This process must be manually repeated by a human expert {for} every phone model, which makes it impractical
for all but the largest companies. The Facebook app alone runs on more than
ten thousand smartphone and tablet models spanning {many} distinct
SoCs~\cite{wu2019facebook}. New phones are released every year, and roughly
two new open foundation models are released every week~\cite{aiindex2024}.
A configuration that performs well on one phone may perform poorly, or fail
entirely, on another. For example, in our measurements, llama.cpp GPU prefill
is 5.4$\times$ faster than CPU prefill on the OnePlus 15, but 7.4$\times$
slower on the OnePlus 8 Pro. Similarly, Qwen3-4B at 8-bit precision {decodes}
12.6--13.8 tokens/s on the OnePlus 15, but {only 0.7--4.8 tokens/s on the
OnePlus 8 Pro.}
 
We ask the following question: given a task specification and a target phone, can we automatically produce a fine-tuned, quantized SLM that meets both the task's accuracy goal and the phone's resource budget?
The developer inputs the specification, which includes any available data sources, the evaluation metric, a verifier for the model's outputs, optionally an accuracy goal, and the app's resource requirements, such as prefill and decode speed, memory, and energy. This is difficult because we cannot know in advance which model will meet these requirements while achieving the best accuracy. Further, we may not know how large a model a task needs until we train it.
 
We present {\it Mobile AI Factory}, an agentic system that builds specialized SLMs. Its core is a loop that curates training data, fine-tunes and evaluates a model, then decides what to try next. An orchestrator LLM chooses one intervention per iteration, while fixed rules handle rollback, budgets, and model scaling. Three components support this loop: a teacher model that generates {synthetic data}, a held-out-data agent that prepares a held-out set before training and shares only aggregate results with the orchestrator, and a phone profiler that benchmarks models across quantization levels, runtimes, and backends once per phone, reusing the results across tasks to identify configurations that meet app requirements.
 
At its core, our factory uses three key techniques.
 
\squishlist
  \item \textbf{Agentic search with rollback and adaptive accuracy goals.} We model each training attempt as a pipeline of dataset, hyperparameters, and learning strategy. At each iteration, the orchestrator chooses one intervention: tuning hyperparameters (LoRA rank, scaling, learning rate, weight decay, and number of epochs), mining more real data (if available) from the task's sources, or having the teacher generate synthetic examples aimed at the failure categories that hurt accuracy most. A contamination check removes any row that also appears in the held-out test set. Since most interventions do not help, any change that lowers accuracy is rolled back to the best model so far, and the orchestrator tries something different. Finally, a fixed accuracy target may be too easy for some tasks and unreachable for others. Instead we set the goal to the teacher model's five-shot score on the task {(zero-shot where five examples do not fit its context)}, with a floor of 0.8. If the factory reaches the goal quickly or with little effort on the current model, the orchestrator can choose to raise it and continue its search. This lets it keep improving on tasks where the goal was set too low.
 
  \item \textbf{Bucketed search from small to large models.} A model's size depends on both its parameter count and its quantization: a 2B model at 4 bits is smaller than a 16-bit 0.8B model. We group all combinations of base model and quantization into $N$ buckets by size, and start with the smallest bucket. Within a bucket, the orchestrator picks which combination to train. If accuracy stops improving before reaching the goal, the factory moves to the next bucket. This avoids two extremes: letting the orchestrator pick one combination from the whole pool, which gives no guarantee of a small model, and training every
  combination, which is too expensive.
 
  \item \textbf{Profiling the target phone.} Whether a model meets the app’s requirements cannot be reliably determined from its specifications; it must be measured on the phone.  The phone
profiler measures prefill and decode speed, peak memory, and energy on the
target phone for all combinations of base model, quantization, runtime
(llama.cpp or MNN), and backend (CPU or GPU). Each phone is profiled once,
and the profile is reused across all tasks. The profile also determines how
to run the final model, since the best runtime and backend differ across
phones, models, and quantization levels. The factory can then search only over models that meet the
app's requirements on that phone, and the model it outputs is far more
likely to run within them.
\squishend
 
We implement Mobile AI Factory using {Claude Sonnet~5} as the orchestrator. Each run uses two NVIDIA L40S GPUs: one trains and scores the model, and the other hosts a local teacher model ({Qwen3.6-35B-A3B}) that generates synthetic data. The model
pool contains nine base models from 135M to 4B parameters, from the {SmolLM2, Gemma~3, Qwen3, and Qwen3.5} families, each at 4-bit, 8-bit, and 16-bit precision, for 27 combinations. We group these into five buckets by size on disk, with boundaries at 300~MB, 750~MB, 1.5~GB, and 2.5~GB. We profile four Android phones, ranging from a current flagship to a budget phone whose GPU the inference engines cannot use due to the lack of OpenCL: the OnePlus 15, Galaxy S23, OnePlus 8 Pro, and Redmi 9A. We evaluate the factory on ten tasks: biomedical entity extraction, fine-grained entity extraction, calendar event extraction, function calling, command parsing, {151-way intent classification}, spam detection, emotion detection, grammatical error correction, and dialogue summarization.
 
Our results are as follows:
\squishlist

\item On an unseen test set, the factory's  small models outperform both
  Claude Sonnet~5 and DeepSeek~v4 Flash, given five examples, on five tasks,
  by 0.024 to 0.342, and match them on three more. The gains are largest where
  the output format is strict: on biomedical entity extraction, a 517\,MB
  model scores 0.828, while Claude scores 0.652 with five examples.

\item With the phone in the loop, the factory returns models whose profiled cost on the phone meets every limit. On a Galaxy S23 whose limits rule out the factory’s best model for call summaries, it returns a model within 0.013 of it.
 
\item Phone measurements cannot be replaced by reasoning. Given every
  specification but no real measurements, {Claude Opus~5.5}, with high reasoning effort enabled,  misses the
  app's requirements in a quarter of easy and moderate cases, and in 60\% of
  hard ones. It misjudges run-time memory and the backend, neither of which
  appears on a spec sheet.
 
\item After the first fine-tune, search adds a median of 8 points on the
  smallest models and 2 points on the largest. On biomedical entity
  extraction, bucketed search comes within 3 points of training all 27
  combinations at a nineteenth of the GPU time. When a goal is reached within
  a few iterations, the orchestrator raises it; in {seven of the nine} runs
  where a raise led to more training, the model improved{, by 3.6 points}.
\squishend
 
{\bf Contributions.}  (1) We present Mobile AI Factory, to our knowledge the first agentic system that builds a fine-tuned, quantized SLM for a given task and a given phone. In contrast to optimizing accuracy on a single model~\cite{pioneer}, our factory searches from small to large models and uses measurements from the target phone. (2) We profile four Android phones across 364 configurations (104 per phone with a GPU, 52 on the Redmi 9A) of model, quantization, runtime, and backend, and show that the best configuration differs across phones, models, and quantization levels. (3) We evaluate the factory on ten tasks and show that, on unseen test sets, its on-device models outperform two five-shot frontier models on five tasks and match them on three more, and that, with the phone in the loop, it returns models whose profiled cost on the phone meets every limit.

\section{Related Work}
 
\vskip 0.05in\noindent{\bf Agentic and Automated Fine-Tuning.} Classical AutoML automates model selection and
hyperparameter search over predefined
spaces~\cite{autoweka,autosklearn,randomsearch,hyperband,bohb,llambo}, but cannot
change the training data. 
MLE-bench~\cite{mlebench} and Kaggle-style agents~\cite{mlestar,aide}
evaluate agents on end-to-end modeling tasks.
AutoML-Agent~\cite{automlagent} proposes multiple LLM agents to go from a task
description to a working
model, but not one optimized to a particular device. Prompt
optimizers~\cite{dspy,opro,gepa} instead improve the prompts given to a
fixed model.
 
Over the past year, researchers have begun using  agents to fine-tune
language models~\cite{ftdojo}. 
PostTrainBench~\cite{posttrainbench} studies whether agents
can post-train base language models under a fixed compute budget.
Pioneer Agent~\cite{pioneer}, the closest work, acquires data, builds evaluation sets, and
trains small language models by jointly searching over data,
hyperparameters, and learning strategy. Other
work uses agents to curate or process training data for a fixed student
model~\cite{agenticdata,llmautodp}, or to generate synthetic data aimed at
the student's weaknesses~\cite{looptool}. Tools such as
Slemify~\cite{slemify} automate a fixed pipeline of synthetic data
generation, fine-tuning, and quantization for models served on CPU servers. 
 
These systems share a key assumption: the model they produce will run on a server, so accuracy is the only objective. On a phone, this assumption breaks in four ways. First, a search that maximizes accuracy favors the largest model. Our factory instead treats the base model and its quantization level as a single choice and searches from small to large, moving to a larger model only when the current one stops improving. Second, a model evaluated at full precision may not be able to run on the phone; 4-bit quantization can significantly lower the accuracy{: on the same weights, MNN's default 4-bit export loses up to 0.19 (\S\ref{sec:motivation})}. {Our} factory  {therefore scores every model after quantization}. Third, the same model can meet the app's requirements on one phone and fail them on another, since the best runtime and backend differ across phones{, and even a frontier model given full specifications picks a configuration that fails the app's limits in 60\% of hard cases (\S\ref{sec:eval-phone})}. Our factory uses a measured profile of the target phone to decide {which models to search and} how the final model runs. Finally, tasks differ widely in difficulty, so a fixed accuracy target is either unreachable or trivial. Our factory sets its goal from a teacher model and raises it when the goal is met easily.
 
\vskip 0.05in\noindent{\bf Hardware-Aware Model Design.} Hardware-aware
neural architecture search includes device cost in the search objective,
either by measuring latency on real phones~\cite{mnasnet} or by estimating
it with predictors and lookup tables~\cite{proxylessnas,fbnet}. Other work
specializes networks for different devices~\cite{once-for-all} or for
microcontrollers~\cite{mcunet}. {nn-Meter~\cite{nnmeter} predicts DNN inference
latency, and \cite{haq,amc}
search quantization and compression policies with hardware feedback.} These
methods design architectures {or compression policies for a given dataset,
rather than adapting pretrained language models to a new task}. In contrast,
we fine-tune pretrained language models for a new task and treat quantization
as part of the search. {We also measure each configuration on the phone
instead of predicting it, since run-time memory and the best backend differ in
ways a predictor or a specification sheet may miss.}
 
Recent work also gives LLM agents feedback from hardware, to write embedded
firmware~\cite{autoembed,embedagent,iotskills,circuitai,embeddedarena} or generate inference
engines~\cite{vibeserve}. These systems optimize
the software around a model, or preserve a model that already performs the
task. Our factory instead builds the task capability itself, for language
models on phones.
 
\vskip 0.05in\noindent{{\bf Adapting Models to Devices.}  NestDNN~\cite{nestdnn} builds
multi-capacity models that trade accuracy for resources at runtime,
LegoDNN~\cite{legodnn} scales models block by block to fit each device, and
AdaptiveNet~\cite{adaptivenet} adapts a deployed model to diverse edge
environments. These systems start from a model that already performs the task
and adjust its size. Our factory starts from a task specification, builds the
task capability through fine-tuning and data, and chooses the model, its
precision, and its runtime from measurements of each phone.}
 
\vskip 0.05in\noindent{\bf On-Device LLM Inference and Measurement.} A
number of engines run quantized language models on phones, including
llama.cpp~\cite{llamacpp}, MNN~\cite{mnn,mnnllm}, MLC-LLM~\cite{mlcllm}, and
ExecuTorch~\cite{executorch}. Research systems further reduce latency using
mobile NPUs~\cite{llmnpu} and activation sparsity~\cite{powerinfer2}, and
MobileLLM~\cite{mobilellm} designs sub-billion-parameter models for phones.
Prior work also measures LLM performance, memory, and energy across phones
and frameworks~\cite{melting,mobileaibench}, profiles models on real devices
in device farms~\cite{qaihub}, and selects a backend for each device at
runtime~\cite{cactus,executorch}. These systems run or measure an existing
model. In contrast, our factory builds the model, and uses per-phone
measurements to decide how it runs on each target phone.
 
\vskip 0.05in\noindent{\bf Training Data and Quantization.} Prior work shows
that data generated by a teacher model lets small models perform well beyond
their size~\cite{selfinstruct,phi3,orca}, and that a small
set of high-quality examples can outperform a larger one~\cite{lima}.  {Small
models can outperform much larger
models~\cite{distillstepbystep,loraland}; our factory automates
this process and targets it at a phone.} Later work targets this data at
the student's failures~\cite{activesynth,looptool}, filters it for
quality~\cite{bansal}, and warns that self-generated data can narrow a
model's outputs~\cite{modelcollapse}. Our factory uses synthesis as an  intervention, anchored to real training examples and {filtered by
programmatic checks} before training. Similarly, models are typically fine-tuned with
LoRA~{\cite{hu2022lora}} and quantized after training~\cite{llmint8,gptq,awq}.
In contrast, our factory makes quantization part of the search, choosing
between a larger model at lower precision and a smaller one at higher
precision, and scores both after quantization. Unlike systems that
fine-tune on the phone~\cite{mobilefinetuner}, our factory runs on a server.

\providecommand{\old}[1]{\textcolor{blue}{\sout{#1}}}
 
\section{Motivation for hardware profiling}
\label{sec:motivation}
 
Deploying a small language model on a {phone} requires choosing an inference engine, {a backend (CPU or GPU), and a quantization}. These choices depend heavily on the phone.
 
\vskip 0.05in\noindent{\bf Measurement setup.} We benchmark 26 of the 27 model variants that Mobile AI Factory searches over: nine base models from 135M to 4B parameters, each at {4, 8 and 16 bits}, except the 16-bit Qwen3.5-4B, which exceeds the RAM of two of our {four} phones. We run them on four phones that span five years: a OnePlus 15 (Snapdragon 8 Elite Gen 5, 16\,GB RAM), a  Galaxy S23 (Snapdragon 8 Gen 2, 8\,GB), a OnePlus 8 Pro (Snapdragon 865, 12\,GB) and a Redmi 9A (MediaTek Helio G25, 4\,GB). Each variant runs on  llama.cpp~\cite{llamacpp} and MNN~\cite{mnn}, on  the CPU and the GPU{; the Redmi 9A's GPU lacks OpenCL, so it runs on the CPU only.}

We follow the protocol of~\cite{XXX-2607.05475}{:} a 256-token prompt{,} then  256 {generated} tokens,  four CPU threads, and three {repetitions}  after one warm-up. Every run starts with the battery temperature at or below 28\,$^\circ$C (32\,$^\circ$C on phones that idle above {it}), with the CPU clocks uncapped, on battery, in airplane mode with Wi-Fi left on for adb, and with the screen off{; the OnePlus 8 Pro stays awake with only its panel off, since it caps its CPU when it sleeps. Energy is the whole phone's battery energy, read from the fuel gauge every 100\,ms, minus idle power. } At 4 bits, the profiles use Q4\_0 for llama.cpp, {since the pinned build's} OpenCL backend has no Q4\_K\_M kernel, and MNN's default export; the factory scores accuracy on the  Q4\_K\_M and HQQ exports of the same weights (\S\ref{sec:scoring}).  We attempted 364 configurations (104 on each phone with a GPU, 52 on the Redmi 9A){,} and 288 ran to completion. Repeated runs  differ by 0.1--5\% within a session (up to {38\%} for MNN's CPU prefill on the {OnePlus 15}){,} while GPU decode throughput varies by 10--20\% between sessions, small next to differences of up to 94$\times$. 
 
\vskip 0.05in\noindent{\bf Key Findings.} {Our findings are as follows:}
 
\squishlist
 
\begin{figure}[t]
  \centering
  \includegraphics[width=0.8\columnwidth]{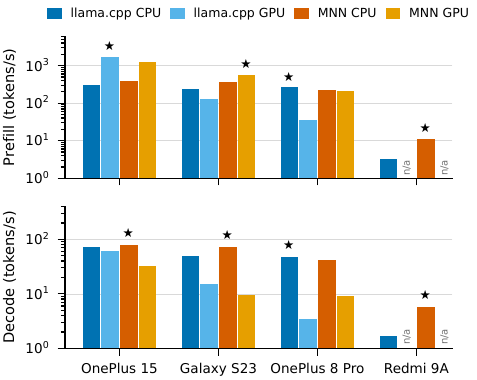}
  \vskip -0.15in
  \caption{Qwen3-0.6B at 4 bits on four phones, two inference engines and two backends (CPU/GPU). \textnormal{The star marks the fastest option on each phone.}}
  \vskip -0.15in
  \label{fig:speed-by-phone}
\end{figure}

\item {\bf The same model runs 42$\times$ slower on one phone than on another.} {Qwen3-0.6B} on llama.cpp's CPU backend at 4 bits generates 72 tokens/s on the OnePlus 15, 50 on the S23, 48 on the OnePlus 8 Pro, and 1.7 on the Redmi 9A { (Fig.~\ref{fig:speed-by-phone})}. Prompt processing spreads further, from 309 tokens/s  to 3.3, a 94$\times$ gap. A  gap between a flagship and a budget phone is expected. {More surprising is how much the three flagships disagree on everything else.}
 
\begin{figure*}[t]
  \centering
    \includegraphics[width=0.9\textwidth]{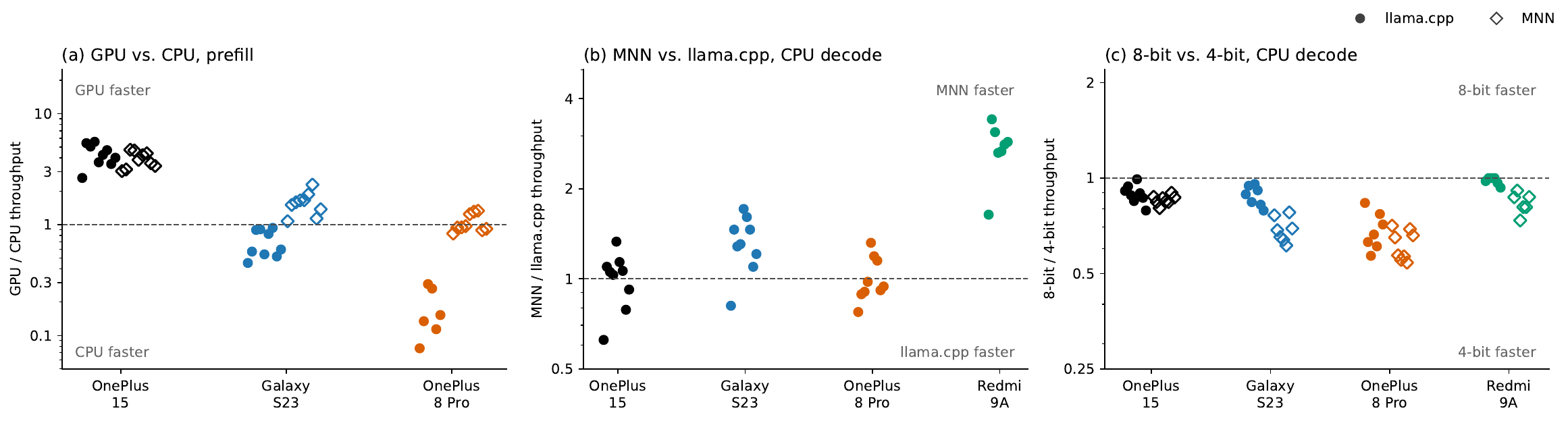}
  \vskip -0.15in
  \caption{\textnormal{Each point is one model; the dashed line marks a tie. (a)~GPU prefill throughput relative to the CPU on the same phone. (b)~MNN decode throughput relative to llama.cpp on the CPU, at 4 bits. (c)~8-bit decode throughput relative to 4-bit on the CPU. Whether a rule holds depends on the phone.}}
  \vskip -0.15in
  \label{fig:rules-flip}
\end{figure*}

\item{\bf The best backend on one phone can be the worst on another.}  For {Qwen3-0.6B} with llama.cpp, {GPU prefill} is 5.4$\times$ faster {than CPU prefill} on the OnePlus 15, 1.7$\times$ slower on the S23, and 7.4$\times$ slower on the OnePlus 8 Pro  (Fig.~\ref{fig:rules-flip}a). A developer tuning on the OnePlus 15 would {put prefill on} the GPU{, and make it} more than seven times slower{ on the OnePlus 8 Pro}. The {best} backend  also {depends on the stage}. On the OnePlus 15, the best GPU configuration prefills a median {3.4$\times$} faster than the best CPU one, while the CPU decodes faster for 14 of the 18 model--precision pairs below 2B parameters.  On the S23,  MNN {decodes SmolLM2-135M} at 224 tokens/s on the CPU but 9  on the GPU, a 25$\times$ {gap}.
 
\item{\bf {Engine} and precision rankings  flip.} MNN's CPU backend  decodes {Qwen3-0.6B} 1.1$\times$ faster than llama.cpp on the OnePlus 15, 1.5$\times$  on the S23 and 3.4$\times$ on the Redmi 9A, but slightly slower on the OnePlus 8 Pro{ (Fig.~\ref{fig:rules-flip}b)}. {Moving from 4 to 8 bits with llama.cpp on the CPU} costs at most 21\% of decode speed on the OnePlus 15 and the S23, but 16--43\% on the OnePlus 8 Pro{ (Fig.~\ref{fig:rules-flip}c)}. On the S23's GPU, {Qwen3-0.6B even decodes faster at 8 bits (26.5 tokens/s) than at 4 (14.9), likely because llama.cpp keeps a 4-bit file's output layer at 6 bits and} the S23's OpenCL kernel for that format is slow.
 
\item{\bf Inference engines can change accuracy.} {Both engines offer 4-bit formats, but} MNN's default export rounds every linear layer to the nearest 4-bit value in blocks of 64 weights{, while llama.cpp's} Q4\_K\_M searches a scale and offset for each block of 32  and keeps the most sensitive layers at 6 bits. {On the same weights, scored on 151-way intent classification (Table~\ref{tab:mnn-acc}), MNN's default drops} a fine-tuned SmolLM2-360M  from 0.767 to 0.659, and {a} zero-shot Qwen3-1.7B  from 0.538 to 0.351. {HQQ} with 32-weight blocks brings six of the eight cells within  noise of llama.cpp or above it. {Since MNN decodes 4-bit models fastest on three of our four phones, a factory} has to score accuracy on the artifact each engine will run, not on a generic quantized checkpoint.
 
\begin{table}[t]
  \centering
  \footnotesize
  \begin{tabular}{lccc}
    \toprule
    Weights & llama.cpp & MNN & MNN \\
            & Q4\_K\_M & default & HQQ-32 \\
    \midrule
    SmolLM2-360M, fine-tuned & 0.767 & 0.659 & 0.747 \\
    Qwen3.5-0.8B, fine-tuned & 0.864 & 0.850 & 0.859 \\
    Qwen3-1.7B, fine-tuned   & 0.886 & 0.851 & 0.875 \\
    Qwen3-4B, fine-tuned     & 0.915 & 0.897 & 0.918 \\
    \midrule
    SmolLM2-360M, zero-shot  & 0.114 & 0.104 & 0.119 \\
    Qwen3.5-0.8B, zero-shot  & 0.372 & 0.303 & 0.320 \\
    Qwen3-1.7B, zero-shot    & 0.538 & 0.351 & 0.557 \\
    Qwen3-4B, zero-shot      & 0.743 & 0.708 & 0.745 \\
    \bottomrule
  \end{tabular}
  \caption{Macro-F1 on CLINC150 for the same weights under three 4-bit recipes. \textnormal{All three are labelled 4-bit. Run-to-run noise is about $\pm$0.015.}}
  \vskip -0.21in
  \label{tab:mnn-acc}
\end{table}
 
\item{\bf Spec sheets do not reveal which configurations will run.} {Of the 364 configurations, 76 did not run. Weight size predicts 35 of these. The other 41 failed for reasons a spec sheet does not show: 20 MNN 16-bit GPU runs crashed, 10 swapped although their weights fit in RAM, 8 hit GPU-driver gaps, and 3 failed to load.} The OnePlus 8 Pro's  driver  lacks OpenCL symbols  the newer llama.cpp  needs, which rules out llama.cpp on the GPU for every {Qwen3.5} model. {RAM is a poor guide too: the} OnePlus 8 Pro has more RAM than the S23, yet  runs {Qwen3-4B} at 8 bits at only 0.7--4.8 tokens/s, against 12.6--13.8 on the OnePlus 15.  File size also underestimates memory. On the S23, llama.cpp's GPU backend holds a median 2.8$\times$ the model's file size, and up to 8.1$\times$ for the smallest model, since it allocates GPU buffers for weights, KV cache and workspace while keeping a CPU copy of the weights.
 
\squishend
 
\begin{table}[t]
\vskip -0.15in
  \centering
  \footnotesize
  \begin{tabular}{llll }
    \toprule
    Phone & Model & Engine, backend & F1 \\
    \midrule
    \multicolumn{4}{l}{\emph{Budget: 1\,s}} \\
    OnePlus 15    & Qwen3.5-0.8B, 4-bit & MNN, GPU         & 0.835 \\
    Galaxy S23    & SmolLM2-360M, 4-bit & MNN, CPU         & 0.744 \\
    OnePlus 8 Pro & SmolLM2-135M, 4-bit & llama.cpp, CPU   & 0.719 \\
    Redmi 9A      & \multicolumn{3}{l}{none (fastest option takes 10\,s)} \\
    \midrule
    \multicolumn{4}{l}{\emph{Budget: 4\,s}} \\
    OnePlus 15    & Qwen3.5-2B, 8-bit   & llama.cpp, GPU   & 0.859 \\
    Galaxy S23    & Qwen3.5-2B, 8-bit   & MNN, GPU         & 0.859 \\
    OnePlus 8 Pro & Qwen3.5-0.8B, 4-bit & MNN, CPU         & 0.835 \\
    Redmi 9A      & \multicolumn{3}{l}{none} \\
    \bottomrule
  \end{tabular}
  \caption{\textnormal{The most accurate deployment of the entity-extraction model for each phone. When several engines and backends reach the same accuracy, we list the fastest. }}
  \vskip -0.2in
  \label{tab:frontier}
\end{table}
 
\vskip 0.05in\noindent{\bf Implications for Mobile AI Factory.} To see what {this means} for a developer, we take the biomedical entity-extraction task and its accuracy for {the 26 profiled variants}, and ask: if a request (a 256-token prompt and a 48-token answer) must finish within a fixed time, what is the most accurate model each phone can run? {Under} a 1\,s budget, each phone needs a different model,  on a different engine and backend, and accuracy ranges from 0.719 to 0.835{ (Table~\ref{tab:frontier})}. Under a 4\,s budget, the OnePlus 15 and the S23 both reach the best accuracy any variant achieves on this task,  through different engines. 
 
{These} measurements shape three decisions in Mobile AI Factory. First, the  configuration cost has to be measured on each phone{;} {the} measurements do not depend on the task, so we run them once per phone and reuse them for every model the factory builds. Second, accuracy has to be measured on the exact artifact the target  engine will run. Third, the right model size cannot be fixed before the phone is known. 

\begin{figure*}[t!]
  \centering
  \includegraphics[width=0.85\textwidth]{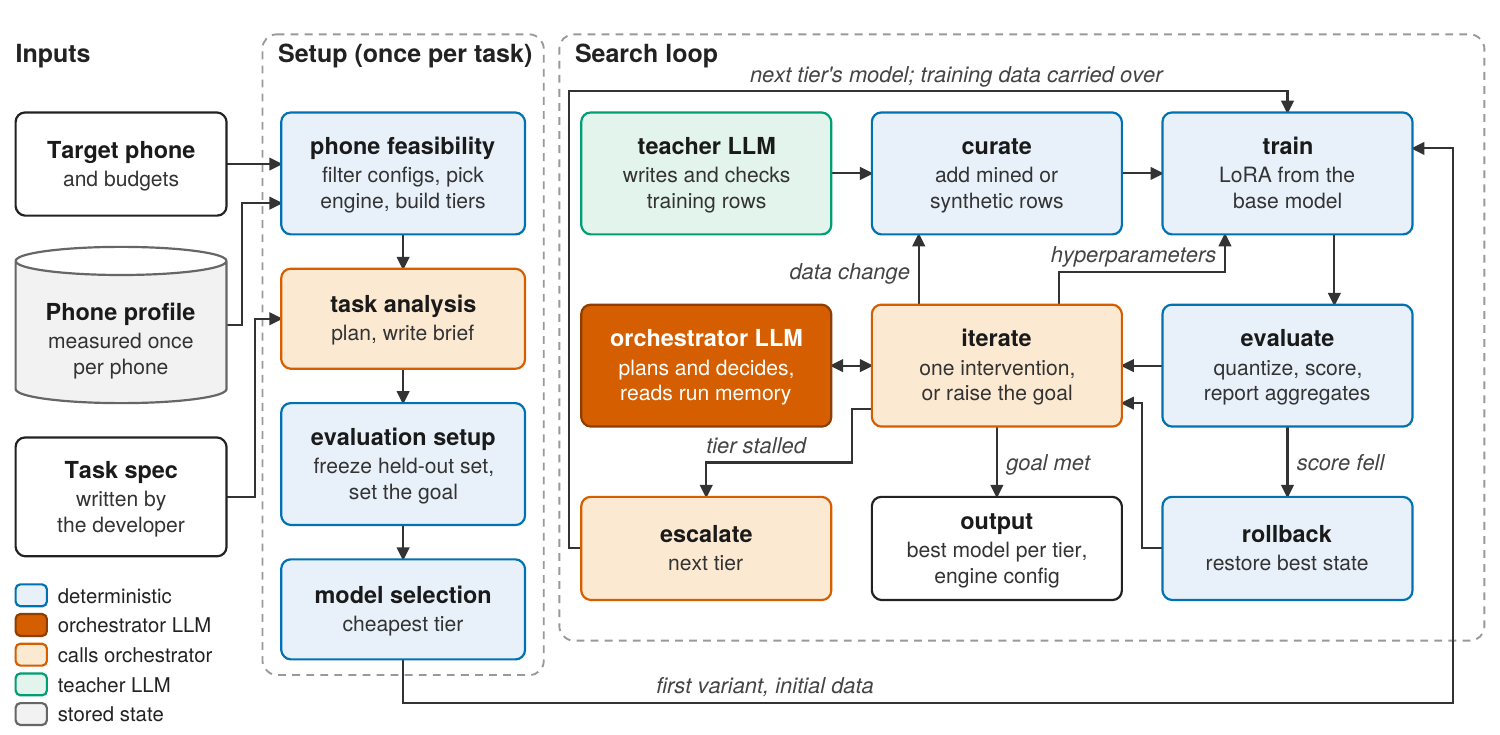}
\vskip -0.15in
  \caption{Mobile AI Factory. \textnormal{ A setup stage reads the phone's measured profile and builds a frozen evaluation set; a loop then curates data, trains, scores the quantized artifact and decides what to do next. {Dark orange is the orchestrator LLM, and light orange nodes call it. Blue} nodes, including rollback and the decision to escalate, are deterministic code. }}
  \vskip -0.15in
  \label{fig:arch}
\end{figure*}
 
\section{Mobile AI Factory Design}
\label{sec:design}
 
Our factory takes two inputs: a task specification and a target phone. It returns a fine-tuned, quantized model for that phone, together with the inference engine and backend. 
 
\subsection{Agentic Architecture Overview}
\label{sec:arch}
 
Fig.~\ref{fig:arch} shows our two-stage system. The setup stage{, run once per task,} looks up the target phone's measured profile, removes every configuration that exceeds the developer's budgets on that phone, and picks the inference {engine and backend for each remaining variant}. The search loop  runs until the goal is met or the budget runs out.
 
{The} orchestrator LLM decides which intervention to try next, which model to try within a tier, and whether to raise the accuracy goal. It does not decide when to stop, when to roll back, or when to move to a larger model{; those are} deterministic rules. We split the roles this way for two reasons. First, a run calls the orchestrator most often when it is stuck, so the rule that notices it is stuck should not depend on the orchestrator. Second, most of the orchestrator’s ideas do not work: in the run with the most iterations (nested intent parsing, 109 evaluations, 104 of them interventions), 88\% of the interventions made the model worse and were rolled back. The rules ensure  steady progress. 

{Each run uses three models: the orchestrator, a local teacher that writes synthetic training data, and the small model being built for the phone.} 
 
\subsection{Setup Stage}
\label{sec:setup}
 
The setup stage finds the configurations the phone can run, writes a brief for the teacher model, and builds the evaluation set.
 
\subsubsection{Phone Profiles and Feasibility}
\label{sec:profiles}
We measure each phone once, using the protocol in \S\ref{sec:motivation}. A configuration is one of the  model variants run on one engine (llama.cpp or MNN) and one backend (CPU or GPU). For each configuration, the profile stores prefill and decode throughput, peak memory, and energy per token; configurations that crash, run out of memory, or are rejected by the GPU driver are stored as failures.
 
The developer gives budgets for a typical request: latency, memory and energy. The factory computes each configuration's latency from its measured throughput and the request's prompt and output lengths, and drops configurations that exceed a budget or failed on this phone. {For each surviving variant, it keeps the engine and backend that decode fastest on that phone.}
 
The factory groups the feasible variants into five tiers by their size on disk, with boundaries at 300\,MB, 750\,MB, 1.5\,GB and 2.5\,GB. A 2B model at 4 bits (1.3\,GB) thus shares the third tier with a 0.6B model at 16 bits (1.4\,GB). Of the 27 variants, 6, 5, 5, 4 and 7 fall in tiers 1 to 5. A phone profile only removes variants; it does not change the tier boundaries.
 
\subsubsection{Task Analysis and the Brief}
\label{sec:brief}
The orchestrator reads a sample of training rows and writes a short brief describing the task, what a correct answer should look like, and the mistakes most likely to make an answer incorrect. {It} never sees the held-out  set. {The brief is included in every} prompt that asks the teacher to generate or check a training row.
 
\subsubsection{Evaluation Setup}
\label{sec:evalsetup}
 
{The} factory draws up to 1,000 rows, balanced across labels for classification tasks, and freezes them {as the held-out set}. The set is scored on every iteration, so we cap it.  {Four firewalls check} for overlap with the held-out set{, by exact match after normalization,} at four points: (1) when a dataset is loaded, where {training rows that repeat a held-out row are dropped and any overlap that remains stops the run}; (2) on every row mined or generated later; (3) again after quality control, which can rewrite rows; and (4) through the label set, which is fixed from the held-out rows so that no later step, including the teacher, can add a label. The trainer also keeps its own validation split for choosing checkpoints, so the held-out set is never seen during training. {The held-out set guides every search decision, so its scores can read slightly high; \S\ref{sec:eval-unseen} measures this on  unseen test sets.}
 
{To label difficulty, the factory runs} the smallest and the largest base models over the held-out set, without fine-tuning. A row is {\em easy} if both answer it correctly, {\em medium} if only the larger one does and {\em hard} if neither does. This measures the gap the tiered search aims to close. A fixed {accuracy} target such as 0.9 may be trivial for some tasks and out of reach for others, so the factory scores the teacher, prompted with five real training examples, on the held-out set through the task's own scorer and sets the goal to that score, with a floor of 0.80 and a cap of 0.99. The run starts in the lowest tier.
 
\subsection{Search Loop}
\label{sec:loop}
 
The loop decides which model to train (\S\ref{sec:ladder}), what to change (\S\ref{sec:interventions}) and whether a change helped (\S\ref{sec:scoring}).
 
\subsubsection{Small-to-Large Search}
\label{sec:ladder} {The} factory starts in the cheapest tier  and moves up only when that tier stops improving.
 
\vskip 0.05in\noindent{\bf Escalation rule.} The factory escalates when the score has not risen by more than 0.02 over the last 15 evaluations, or after 30 evaluations in one tier without reaching the goal. Both counts include rolled-back attempts. On escalation, the orchestrator picks one variant in the next tier. The run keeps its training data, and the rest of the search starts over.
 
\vskip 0.05in\noindent{\bf Increasing the goal.} When the run reaches its goal, the factory stores the result {and asks} the orchestrator whether to raise the goal, by at least 0.005 above the highest goal so far and to at most 0.99. A raised goal sends the run back into the loop. The orchestrator may lower the goal when failures look like a model size limit, but {only after a raise, and never below the starting goal}.
 
\vskip 0.05in\noindent{\bf Output.} The run returns the best model in each tier it visited, with its engine and backend. Since the search climbs from the cheapest tier, the developer reads off the cheapest one that meets their bar. If no tier does, the run returns its best model and why it stopped: tiers, steps or time ran out.
 
\subsubsection{The Orchestrator and Its Interventions}
\label{sec:interventions}
 
In each iteration, the orchestrator proposes exactly one change.
 
\vskip 0.05in\noindent{\bf Input and output.} The orchestrator  sees  the {evaluator's} report{ (}the overall score, accuracy on easy, medium and hard rows, and the most common failure categories from the task's own scorer, such as wrong entity boundaries or a call to an undeclared function{)}{, and} the run memory, built from the search graph, which lists every attempt, whether it was kept, and what has failed since the last improvement. It returns a JSON object, naming the change, its parameters and  {a} hypothesis. The factory validates it, and replaces a hyperparameter setting already tried on the same data with an untried {neighbor}.
 
\vskip 0.05in\noindent{\bf Hyperparameters.} A hyperparameter intervention holds the data fixed and adjusts up to five values: LoRA rank, the ratio of $\alpha$ to rank, weight decay, learning rate and epochs. Batch shape follows from the GPU  and dropout is fixed at zero.
 
\vskip 0.05in\noindent{\bf Data.} A data change adds rows and keeps the best hyperparameters so far. Rows come from two sources. {\em Mining} first reads further into the task's own sources, and searches the web for new datasets only when those are exhausted, giving up after two empty rounds. {\em Targeted synthesis} has the teacher write 25--600 rows per failure category, in proportion to its failure count, and drops a category whose count does not fall. Synthesis passes three gates. (1)~The teacher must score at least 0.80 on the task with five real examples, the way synthesis prompts it, or synthesis is off for the run{; our main runs bypass this gate on seven tasks (\S\ref{sec:eval-setup})}. {Examples matter: on} biomedical entity extraction the teacher scores 0.13 without them and 0.32 with five. (2)~Each row passes a programmatic check where one exists{, for example that every extracted entity appears verbatim in its text}. (3)~The teacher {can} check each row against the brief (\S\ref{sec:brief}), using only real examples as references{; our runs turn this check off where a strict programmatic check exists and, on the other tasks, record its verdicts without dropping rows}. The training set only grows; rows leave only through quality control{,} the firewall{, or the rollback of the change that added them}.
 
\vskip 0.05in\noindent{\bf Rollbacks.}  Three mechanisms limit each failure to one iteration. Training always restarts from the base model, so rollback exactly restores the previous behavior. Control then returns to the orchestrator, which must try a different change rather than repeat the same regression. A watchdog stops runs that stop making progress, like three consecutive synthesis rounds that add nothing or a training set that shrinks by 25\% (or 1,000 rows) in one step. We require repeated or large failures so the watchdog does not trigger on noise{; we added it} after one run spent eight hours on eight data changes that added no examples.
 
\begin{table*}[t]
\centering
\small
\setlength{\tabcolsep}{3.0pt}
\begin{tabular}{llccccclr}
\toprule
 & & Claude & DeepSeek & & & & & \\
Task & Metric & 0-shot / 5-shot & 0-shot / 5-shot & Untrained & First FT & Factory & Factory model & MB \\
\midrule
\multicolumn{9}{l}{\emph{Output-format bound}} \\
\midrule
BC5CDR        & span F1          & 0.165 / 0.606 & 0.128 / 0.508 & 0.000 & 0.705 & \textbf{0.830} & Qwen3.5-0.8B Q4 & 517 \\
Calendar JSON & arg.\ match      & 0.514 / 0.634 & 0.505 / 0.669 & 0.036 & 0.535 & \textbf{0.811} & Qwen3.5-0.8B Q4 & 517 \\
TOPv2         & exact match      & 0.241 / 0.242 & 0.145 / 0.085 & 0.003 & 0.534 & \textbf{0.599} & Qwen3-4B Q8 & 4,082 \\
MultiCoNER    & micro F1         & 0.491 / 0.587 & 0.428 / 0.521 & 0.284 & 0.604 & \textbf{0.653} & Qwen3-4B Q8 & 4,082 \\
BFCL          & arg.\ match      & 0.879 / 0.875 & \textbf{0.896} / 0.876 & 0.818 & 0.834 & 0.852 & Qwen3.5-4B Q4 & 2,655 \\
\midrule
\multicolumn{9}{l}{\emph{Label not directly readable from the input}} \\
\midrule
CLINC150      & macro F1         & 0.921 / \textbf{0.933} & 0.907 / 0.904 & 0.114 & 0.770 & 0.884 & SmolLM2-360M Q4 & 258 \\
SMS Spam      & minority F1      & 0.954 / \textbf{0.961} & 0.856 / 0.832 & 0.235 & 0.878 & 0.913 & SmolLM2-360M Q4 & 258 \\
GoEmotions    & Ekman F1         & 0.506 / 0.481 & 0.448 / 0.436 & 0.378 & 0.598 & \textbf{0.598} & Qwen3-4B Q8 & 4,082 \\
BEA-19        & ERRANT F$_{0.5}$ & 0.476 / 0.506 & 0.518 / 0.528 & 0.435 & 0.536 & \textbf{0.565} & Qwen3-4B Q8 & 4,082 \\
\midrule
\multicolumn{9}{l}{\emph{In distribution}} \\
\midrule
DialogSum     & ROUGE-L          & 0.334 / 0.424 & 0.327 / 0.374 & 0.272 & 0.482 & \textbf{0.482} & Qwen3-4B Q8 & 4,082 \\
\midrule
\multicolumn{9}{l}{\emph{Goal pinned at 0.99}} \\
\midrule
CLINC150      & macro F1         & 0.921 / 0.933 & 0.907 / 0.904 & 0.766 & 0.901 & \textbf{0.937} & Qwen3-4B Q8 & 4,082 \\
SMS Spam      & minority F1      & 0.954 / \textbf{0.961} & 0.856 / 0.832 & 0.000 & 0.707 & \textbf{0.961} & Qwen3-1.7B Q4 & 1,223 \\
\bottomrule
\end{tabular}
\caption{Factory models against frontier models, on each task's own scorer and the same frozen held-out set. \textnormal{ Frontier models are given no examples (0-shot) or five worked examples (5-shot). The three student columns describe the factory's best model: untrained, after its first fine-tune, and after search. Q4 and Q8 are 4- and 8-bit weights; MB is the model's file size. Bold marks the best score in each row; on the last two rows, the factory and Claude are within our 0.015 noise floor.  GoEmotions is scored as macro F1 over its Ekman emotion groups.}}
\vskip -0.15in
\label{tab:main}
\end{table*}
 
\subsubsection{Scoring Each Iteration}
\label{sec:scoring}
 
The setup stage freezes a held-out evaluation set that training cannot access (\S\ref{sec:evalsetup}). The loop {adds} three  rules.
 
\squishlist
 
\item Each iteration merges the adapter, exports {4- and 8-bit models} to {their} deployment format, loads {them} in the target engine, and scores {them} on the held-out set{; 16-bit models are scored in Transformers}. For MNN, we use 4-bit HQQ with 32-weight blocks because its default quantization loses up to 0.19 accuracy (\S\ref{sec:motivation}). 
 
\item The {evaluator} returns scores, failure counts, and the fraction of parseable outputs, but not the evaluation rows. This {separates} wrong answers from malformed outputs. Any decision that quotes a held-out row is rejected.
 
\item The base model is the first candidate, and any fine-tuned model that scores worse is discarded{, so the loop can conclude} that fine-tuning does not help. {A} failed evaluation is recorded as missing, not zero. An early run used zero and incorrectly credited fine-tuning with a gain of 0.85.
 
\squishend
 
\subsection{Implementation Details}
\label{sec:impl}
 
Mobile AI Factory's loop is a LangGraph state machine. The orchestrator is {Claude Sonnet~5}, called through its API, and the teacher is Qwen3.6-35B-A3B, served locally with vLLM. Training uses LoRA adapters on 4-bit base models, through Unsloth, PEFT and Transformers, with loss computed on the answer tokens only and the checkpoint chosen on a validation split. Each run uses two NVIDIA L40S GPUs on a SLURM cluster: one trains and scores the model, and the other serves the teacher. Scoring exports each {4- or 8-bit} model with llama.cpp or MNN's own tools and runs it in that engine on the same GPU. A run can take several days, longer than a cluster job is allowed to run{, so the} run state is checkpointed to SQLite after every step, and a job nearing its time limit checkpoints and requeues itself.

\section{Evaluation}
\label{sec:eval}

\subsection{Setup}
\label{sec:eval-setup}

 \noindent{\bf Tasks.} We use ten tasks{. Five} are bound by their output format{:} biomedical entity extraction (BC5CDR), calendar extraction to JSON, nested intent parsing (TOPv2), entity extraction over 33 fine-grained types (MultiCoNER) and function calling (BFCL). In four, the label cannot be read off the input: 151-way intent classification (CLINC150), spam detection, 28-way emotion classification (GoEmotions) and grammatical error correction (BEA-19). In the last, dialogue summarization (DialogSum), the base model can already do the task and fine-tuning mainly teaches the answer style. Each task starts from 2,600--5,000 real training rows. Its held-out set has 1,000 rows, except for DialogSum (500), Calendar JSON (535) and MultiCoNER (871).

\noindent{\bf Factory configuration.} The model pool holds the 27 variants from \S\ref{sec:profiles}: nine base models from 135M to 4B parameters, each at 4, 8 and 16 bits, from three families:  {SmolLM2} (135M, 360M), {Gemma~3} (270M) and {Qwen} ({Qwen3} at 0.6B, 1.7B and 4B; {Qwen3.5} at 0.8B, 2B and 4B). Every run in this section uses the full pool and the five size tiers of \S\ref{sec:profiles}, and scores every {4- and 8-bit} model as a llama.cpp file (Q4\_K\_M or Q8\_0) on the training GPU{; 16-bit models are scored in Transformers}. \S\ref{sec:eval-phone} then runs the factory {with} the phone profiles. The orchestrator is Claude Sonnet~5 and the teacher is Qwen3.6-35B-A3B. Every run starts from LoRA rank 16, $\alpha$ 32, learning rate $2{\times}10^{-4}$, weight decay 0.01 and 3 epochs.

\noindent{\bf Goal settings.} Every task runs with the factory's default goal, the teacher's five-shot score with a floor of 0.80 (\S\ref{sec:evalsetup}){; for intent classification, whose five examples do not fit the teacher's context, its zero-shot score}. To show how {far} the factory can go when a developer asks for more, we also ran intent classification (CLINC150) and spam detection with the goal pinned at 0.99, so that the search continues through every tier. Table~\ref{tab:main} reports these runs in their own block{. The per-tier analyses below (\S\ref{sec:eval-search}--\S\ref{sec:eval-orch-cost}) use the ten {\em main runs}: the default-goal runs for eight tasks and these two pinned runs, which cover the full ladder}. We also enabled synthesis on every task, including seven where the teacher scored below the factory's 0.80 gate (\S\ref{sec:interventions}), so we can measure its contribution on all ten (\S\ref{sec:eval-interventions}).

\noindent{\bf Baselines.} For every task we score Claude Sonnet~5, DeepSeek v4-Flash, the teacher, and the untrained student. We called both APIs with reasoning disabled, via their batch interfaces.

For each tier of a run, we report three scores: the untrained model, its {\em first fine-tune} on the training data the run holds when it reaches that tier, and its {\em best} score after search. The {\em search gain} is best minus first fine-tune: what the iterative search bought on top of plain fine-tuning. A run's {\em best model} is the highest-scoring model across all tiers.

{Since a run can take days, each} configuration is a single run. All scores are on the held-out set. Two {models trained identically on the same data} differ by up to 0.015 on this set, so we treat smaller differences as ties. The effects we report are  larger: on the seven tasks the factory wins under its default goal, its margin over the better frontier model is {0.037--0.357, over the better of zero- and five-shot}.

Below, we  evaluate the factory first phone-agnostic (\xref{sec:eval-frontier}) and then phone-aware (\xref{sec:eval-phone}).


\subsection{Evaluating the Factory's SLMs}
\label{sec:eval-frontier}

Table~\ref{tab:main} compares the factory's best models with two frontier
models, given no examples or five. Against the better of the two, the factory
wins seven tasks under the default goal, by 0.037 to 0.357. With the goal
pinned at 0.99, it matches Claude on intent classification and spam
detection, within our noise floor. On unseen test sets the factory never saw (\S\ref{sec:eval-unseen}), 5 of these wins
hold and 3 tasks are ties.

 Our local teacher, which sets each run's goal, behaves the same way.
Without and with five examples, it scores 0.132 and 0.321 on biomedical
extraction, 0.204 and 0.624 on calendar extraction, 0.023 and 0.018 on nested
parsing, 0.431 and 0.538 on fine-grained entity extraction, 0.866 and 0.856 on
function calling, 0.864 and 0.850 on spam detection, 0.423 and 0.432 on
emotion classification, 0.501 and 0.509 on grammar correction, and 0.284 and
0.339 on dialogue summarization. On intent classification it scores 0.877.

On tasks frontier models already do well, the margins are small.  With the default goal, the factory stops as soon as a model meets the goal, so these runs end early{, on the smallest tier with a 258\,MB model: intent classification at 0.884 after 14 iterations and 9.6 hours, and spam detection at 0.913 after five iterations and 0.7 hours, both below Claude. With the goal set to  0.99, the search  continues, and the factory matches Claude}. On these tasks, matching a frontier model takes a 1.7B or 4B model, and a search that is allowed to keep going. {The goal, not the factory, sets this ceiling.}



On function calling (BFCL), the untrained 4B model scores 0.818, and fine-tuning and search together add 0.034, of which the search after the first fine-tune adds 0.018. It ends 0.027 below Claude and 0.044 below DeepSeek. Function calling is a format models are trained heavily on, so there is little left for fine-tuning to teach.


\subsubsection{Decomposing the Gains}
\label{sec:eval-search}
{Table~\ref{tab:main}} reports, for the best model on each task, the untrained score, the score after the first fine-tune, and the final score. The first fine-tune accounts for a median of 82\% of the total improvement over the untrained model. On biomedical entity extraction,  it raises the score from 0.000 to 0.705, {and} the  search adds another 0.125. On emotion classification and dialogue summarization, the first fine-tune is already the best result. When a run moves to a larger model, it carries over the training data collected so far, so the larger model's first fine-tune already includes  data added during the search at smaller {tiers}.

\begin{figure}[t]
  \centering
  \includegraphics[width=0.75\columnwidth]{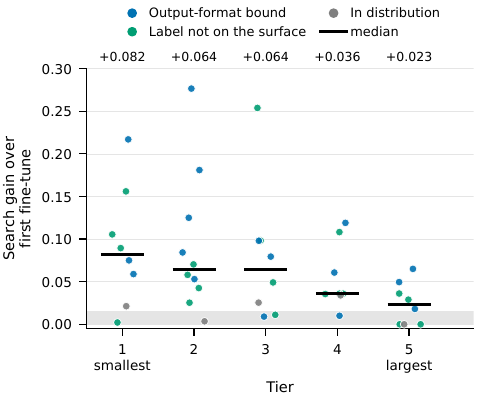}
  \vskip -0.18in
  \caption{What search adds over each tier's first fine-tune. \textnormal{Each dot is one tier of one task's run. The grey band is the 0.015 improvement threshold.}}
  \vskip -0.15in
  \label{fig:search}
\end{figure}

To measure search on its own, we compare each tier's best score with that tier's first fine-tune, across {the ten main runs}. This gives 42 pairs{, after excluding}  two tiers whose 270M model produced parseable output on fewer than 20\% of rows. Search improved on the first fine-tune by more than 0.015 in 34 of them, and by a median of 0.051. The gain shrinks as the model grows (Fig.~\ref{fig:search}). On the smallest tier, search adds a median of 0.082 and closes 21\% of the remaining gap to a perfect score{; on} the largest, it adds 0.023 and closes 9\%.


 \subsubsection{Contribution of Each Intervention}
\label{sec:eval-interventions}
 
\begin{figure}[t]
  \centering
  \includegraphics[width=0.85\columnwidth]{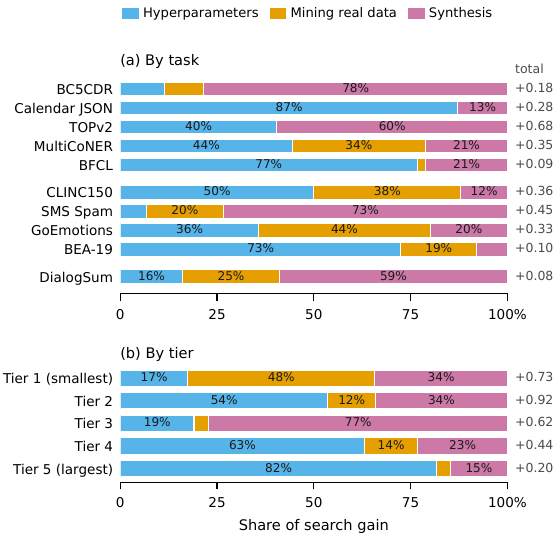}
  \vskip -0.15in
  \caption{Search gains split across  interventions, (a) per task and (b) per tier. \textnormal{Each bar sums the gains of the kept steps of each kind; the number on the right is the total. }}
  \vskip -0.15in
  \label{fig:attr}
\end{figure}
 

  Fig.~\ref{fig:attr} shows that across all tiers, hyperparameter changes carry 41\% of search gains, synthesis 40\% and mining 19\%. One tier inflates synthesis: on spam detection, synthetic rows repaired a 1.7B model {whose first fine-tune had collapsed to 0.707}. Without it, synthesis carries 35\%. {Which intervention matters depends on the task (Fig.~\ref{fig:attr}a): synthesis carries most of the gain on four tasks, hyperparameters on three, and all three contribute on the rest.} 
 

 Mining carries 48\% of the gain on the smallest tier, synthesis 77\% on the third, and hyperparameters 82\% on the largest (Fig.~\ref{fig:attr}b). {This does not mean small models need data and large ones need tuning.} First, mining runs out early, so later tiers fall back on the other two. Second,  training data  carried to {a} larger model improves {its} first fine-tune{, which falls} outside the search. Third, a gain goes to whichever step happened to be kept, {so} a different order could split the same total differently.


On seven tasks the teacher scored below 0.80{, which by default turns} synthesis off  (\S\ref{sec:interventions}). {We allowed synthesis on those runs anyway. They} account for 67\% of all the gain credited to synthesis, mostly on nested parsing and biomedical extraction. The same runs suggest the gate is too strict. It tests whether the teacher can do the task, but the teacher's job is to write training rows. On nested parsing, the teacher scores {0.02} with five examples{, yet} 90\% of the rows it writes pass the task's check, and synthesis carried 60\% of that task's search gain. Writing a correct example is easier than solving a new one{, so a better gate would test the teacher as a writer}.


\subsubsection{Tiered Search versus Baselines}
\label{sec:eval-ladder}
 
\begin{figure}[t]
  \centering
  \includegraphics[width=0.9\columnwidth]{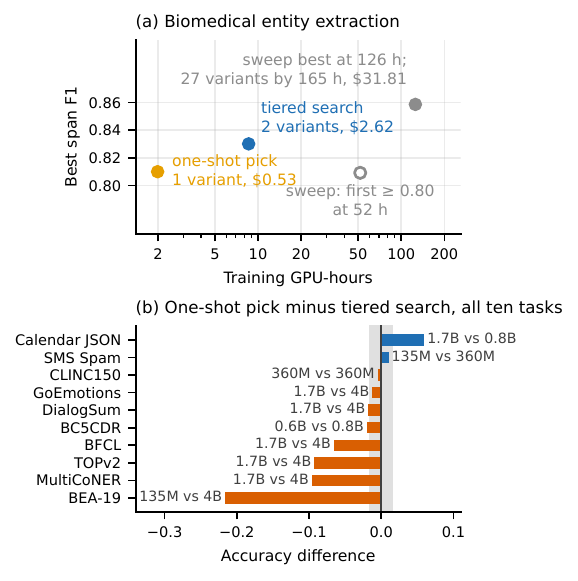}
  \vskip -0.15in
  \caption{(a) Best score against GPU-hours for three search strategies on biomedical entity extraction, with orchestrator spend. (b) One-shot pick compared against the best model from the tiered search on each task, labelled with [picked model vs tiered winner]. The grey band is the noise floor.}
  \vskip -0.15in
  \label{fig:tiered}
\end{figure}


We compare the tiered search with two baselines. An {\em exhaustive sweep} trains all 27 variants with the same loop; we ran it on biomedical entity extraction only, given the compute cost. A {\em one-shot pick} has the orchestrator choose the cheapest variant it expects to reach the goal, and searches only that model; we ran it on all ten tasks. We report GPU-hours on the training GPU and orchestrator spend{; the teacher's GPU is counted in \S\ref{sec:eval-orch-cost}}.


The exhaustive sweep took 165 GPU-hours and \$31.81, and first reached the 0.80 goal after 52 hours (Fig.~\ref{fig:tiered}a). The tiered search reached 0.830 in 8.6 hours for \$2.62{, a nineteenth of the GPU time}. The sweep's best model, a 2B at 8 bits, scored 0.859. But the tiered search stopped as soon as it met its goal, on a 0.8B model whose best in the sweep was 0.835.
Across ten tasks{ with the default goal}, the one-shot pick  used 60 GPU-hours and \$11.68, {against the tiered search's 286 hours and \$45.39. In return, the tiered search scored higher on 8 of 10 tasks, by more than noise on six, and by 0.066--0.216 on four, a median gain of 0.019. Its winning model is a 4B on six tasks; the one-shot pick never chose one} (Fig.~\ref{fig:tiered}b).

\subsubsection{Raising the Goal}
\label{sec:eval-raise}

\begin{figure}[t]
  \centering
  \includegraphics[width=0.86\columnwidth]{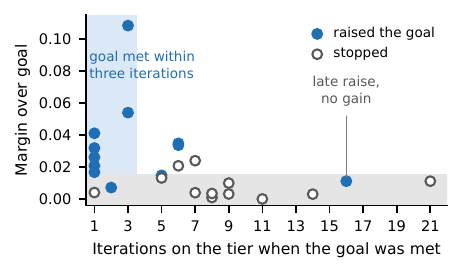}
  \vskip -0.15in
  \caption{Every decision to raise the goal or stop, by how many iterations the tier had taken to meet the goal and by how much it beat it. The grey band is the noise floor, set at 0.015.}
  \vskip -0.18in
  \label{fig:raise}
\end{figure}

When a run meets its goal, the orchestrator can raise it. Fig.~\ref{fig:raise} shows that the orchestrator used both speed and margin. It raised the goal {8 of 9} times when {the goal was} reached within three iterations, but stopped {9 of 10} times after six iterations. In {nine} runs, a raise led to further training{, and seven} of those runs ended higher, by {0.036 on average}. On runs that stayed on one model, the best score arrived within 1--7 evaluations of the raise. All {nine} extensions together took {14.6} hours on the training GPU. In two further runs, the raised goal was already below the score just reached, so the run finished at once. Only once did a raise change the model{: in} an earlier run of biomedical extraction, the higher goal moved the run from a 0.8B to a 2B, for a gain of 0.020.
{The gains} come from {three tasks: spam detection, calendar extraction and biomedical extraction}. In the other  main runs, the search kept climbing toward the goal through the largest tier (on two by design, with the goal pinned at 0.99), so raising the goal never came up.

 \subsubsection{Cost of the Orchestrator}
\label{sec:eval-orch-cost}


Across the ten main runs{, the orchestrator} cost \$57.37, or \$2.62--8.46 per task. Nearly all of it, 98\%, goes to the decision made each iteration{, at} a median of \$0.064. The cost is the same across tasks and tiers, since the prompt summarizes the run in the same way whatever model it trains{, so a run's bill} grows with its number of iterations, not with model size. This is  small next to training. Each run held two GPUs, one for training and one for the teacher, for 663 GPU-hours in total. 

\subsection{Phone-Aware Factory Deployment}
\label{sec:eval-phone}


We evaluate phone-aware deployment using profiles from four phones: a OnePlus 15, Galaxy S23, OnePlus 8 Pro, and Redmi 9A.  On the one phone shared with prior work~\cite{XXX-2607.05475}, our measurements match within  10\% for GPU prefill and {within} a median of 5\% for decode energy.

\begin{figure*}[t]
  \centering
  \includegraphics[width=0.8\textwidth]{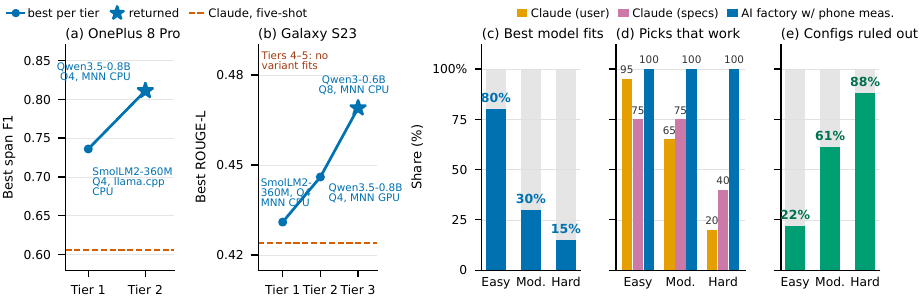}
  \vskip -0.15in
  \caption{Phone-aware factory deployment. \textnormal{(a, b) Two representative end-to-end runs: biomedical entity extraction on a OnePlus 8 Pro (moderate requirement) and call summaries on a Galaxy S23 (hard requirement); labels give each tier's model, precision, engine and backend. (c--e) Across 60 requirements: (c) how often the factory's most accurate model for the task meets the requirement on the phone; (d) how often a configuration picked without measurements meets the requirement. The measured-selection bar is a reference: it meets every requirement by construction; (e) the share of the phone's configurations that measurements rule out before any training.}}
  \vskip -0.15in
  \label{fig:phone}
\end{figure*}

We deploy our ten tasks in {several} settings (\xref{app:settings}). Each deployment specifies limits on memory, latency, throughput and energy. The deployment type sets memory and latency,  the task determines required prefill and decode speeds{, and} an allotted share of daily battery use sets the per-request energy budget. Across four phones, this gives 152 requirements. We classify a requirement as {\em easy} if at least 60\% of configurations satisfy it, {\em moderate} if 30--60\% do, and {\em hard} if fewer than 30\% do{, and study 20 of each}.


 

Fig.~\ref{fig:phone}a,b shows two representative runs, with the phone in the loop.  Fig.~\ref{fig:phone}a shows biomedical entity extraction on a OnePlus 8 Pro. The requirement allows 5,800\,MB of memory, needs at least 5 tokens/s of decode and 3 tokens/s of prefill, and caps each request at 15.7\,J. The phone's profile kept 14 of the 27 variants.  The run took 8.2 hours and \$3.05 of API calls. 
Fig.~\ref{fig:phone}b builds a call summarizer for a Galaxy S23. {The requirement allows} at most 1,024\,MB, {needs} 350 tokens/s of prefill {to start answering quickly}, and {allows} 90\,J per request. Only 11 of the 27 variants meet this. Without the phone, the factory's best model is a 4B at 8 bits, at 0.482 ROUGE-L. It needs 4,082\,MB and cannot run here. With the phone, the factory searched only the 11 variants that fit and returned Qwen3-0.6B at 8 bits, on MNN's CPU backend, at 0.469. That is 0.013 below the model that does not fit, inside our noise floor.  The run took 33.2 hours and {\$2.38}.



This experiment also asks whether a configuration can be chosen without measuring the phone. Each requirement is constructed so that at least one configuration meets it, and every pick, from either method, is checked against the same measured profile. A method that selects from measurements meets every requirement by design; we include it as a reference that shows a feasible configuration exists. The question is how often a selection made without measurements finds one.

Comparing all the phone runs with the phone-agnostic runs, the factory's most accurate model meets 16 of 20 easy requirements, but only 6 moderate and 3 hard ones (Fig.~\ref{fig:phone}(c)); the rest require a smaller model. The best configuration also depends on the phone. Of 18 deployments tested on multiple phones, 15 require {different configurations on different phones}. The same task  uses a 4B 4-bit model on the S23 GPU, a 4B 8-bit model on the OnePlus 15 CPU, and a 2B model on the Redmi 9A. 

We gave Claude Opus 5.5 with high reasoning effort, a stronger model than the factory's orchestrator, the same 60 requirements under two settings. In one, it received all information available without measurement: phone specs, model size and precision, engines, backends, and the four limits. In the other, it saw only the phone name and a plain-language app description. Given the specifications, Claude Opus 5.5 picked a configuration that misses the requirement in 25\% of easy, 25\% of moderate and 60\% of hard cases; given only an app description, in 5\%, 35\% and 80\% (Fig.~\ref{fig:phone}(d)).

\begin{table}[t!]
\centering
\footnotesize
\setlength{\tabcolsep}{2.5pt}
\begin{tabular}{lrr@{\hspace{0.8em}}lrr}
\toprule
Task & CPU & GPU & Task & CPU & GPU \\
\midrule
BC5CDR     & $-$.0004 & $-$.0017 & CLINC150   & $-$.0004 & $-$.0004 \\
Calendar   &    .0000 & $-$.0019 & SMS Spam   & $-$.0160 & $-$.0119 \\
TOPv2      &    .0000 & $+$.0030 & GoEmo.     & $-$.0062 & $-$.0079 \\
MultiCoN. & $+$.0005 & $-$.0005 & BEA-19     & $-$.0007 & $+$.0015 \\
BFCL       & $+$.0010 & $+$.0010 & DialogSum  & $-$.0014 & $-$.0022 \\
\bottomrule
\end{tabular}
\caption{\textnormal{Accuracy of each factory model on the phone's CPU and GPU, minus its accuracy on the server.}}
\vskip -0.17in
\label{tab:ondevice}
\end{table}



Given the spec sheet, Claude underestimated run-time memory and often misjudged prefill speed, which depends on the backend (\S\ref{sec:motivation}). It chose the right inference engine in 92\% of  cases, since MNN was usually best, but the right CPU/GPU backend in only 75\%. In 11 {of 60} cases, the right model at the right precision failed because Claude chose the wrong backend (8 cases) or engine (3 cases). With only a user-style description, it did {better on easy requirements but worse on moderate and hard ones: 15 of its 16 misses on hard requirements chose a larger model or a higher precision than fits, while on easy ones it often underused the phone}. In 10 successful cases, the phone could have run a model with 2--5.6$\times$ more parameters.


Before training anything, the measurements rule out a median of 22\% of a phone's configurations for easy requirements, 61\% for moderate ones and 88\% for hard ones (Fig.~\ref{fig:phone}e). On hard requirements, the largest model that fits is 0.8B or smaller in 19 of 20 cases, so  the factory never needs to train a model of 1.7B parameters or more. The tighter the requirement, the more the measurements save  training time.



Finally, the factory scores every model on a server GPU. While it uses the same engine as the phone, the hardware and the kernels are different. So as a final step, we copy {a factory-trained checkpoint for each task} to a OnePlus 15,  and score it there on the same held-out rows, once on the phone's CPU and once on its GPU. {On the GPU, every} transformer layer and the KV cache run on the GPU;  token {embedding lookup,}  tokenization and sampling still {run} on the CPU. Table~\ref{tab:ondevice} shows the difference from the server. On the CPU, eight tasks differ by at most 0.0014, and two of them give the same score. On the GPU, seven differ by at most 0.0022. The largest differences on both are on spam detection, 0.016 on the CPU and 0.012 on the GPU, at our noise floor, and on emotion classification, 0.006 and 0.008. Spam detection {moves the most} because its score is the F1 of the spam class alone, so a few messages near the decision boundary can change it. The phone's CPU and GPU also agree with each other, to within 0.0041 on every task.
\section{Discussion and Limitations}
\label{sec:discussion}
 
 A developer still  provides the task specification: data sources, the metric, and a verifier for model outputs. Our results suggest that the verifier matters most. Rejecting malformed outputs made synthetic data usable, while tasks where few generated rows survived quality control saw little benefit from synthesis. {Building this harness is the most hand-engineered part of the pipeline. An agent that decomposes a task into its type, such as classification, extraction or generation, and assembles a matching scorer and verifier would make the factory far more autonomous.} {So would} an early check of synthetic-data survival rates.
 
We {also} restrict the factory to a predefined set of  {decoder} models{ from 135M to 4B}. This is practical when developers prefer specific model vendors, but the restriction is not fundamental. The factory could instead download candidate base models on demand{, and once a task is known to be classification, add encoder models of 1--50M parameters, far smaller than any model in our pool. Training is LoRA fine-tuning only; preference optimization, reinforcement learning or distillation from a larger model are natural additions. Synthesis writes a few hundred rows per round from a local 35B teacher. This keeps each round cheap and targeted, but may cap what synthesis can add; larger rounds from a stronger teacher are worth exploring}.

Our phone profiles cover CPUs and GPUs. Recent work shows that mobile NPUs can process prompts an order of magnitude faster while offering little decode advantage over CPUs~\cite{XXX-2607.05475}. Adding NPUs would require only extending the hardware profile, and would make backend selection even more important. {We also measure each configuration alone, from a cool start, at one prompt length, with four threads, and with the phone otherwise idle. A deployed model shares the phone with other apps, other models and changing thermal state, so sustained and concurrent use, and longer prompts, may change which configurations meet a requirement.} Profiling a phone is slow, largely because it must cool between runs, but this cost is paid once and shared across tasks. Future work could measure only a subset of configurations on a new phone and predict the rest, while validating those predictions before the factory relies on them.


\section{Conclusion}
\label{sec:conclusion}
 
We present Mobile AI Factory, an agentic system that builds a fine-tuned, quantized small language model for a given task and  phone.   We showed that the right model and configuration change from phone to phone,  and that even a frontier model cannot make this choice without measurements. 

Taking a step back, ML and agent researchers mainly optimize metrics and rarely think about mobile devices. Mobile systems researchers, on the other hand, build inference engines and take the model as given. Mobile
AI Factory suggests a near-future in which agents build many phone-specific small,
specialized models, and rebuild them as
new phones and open base models are released. The same approach could extend
beyond phones to wearables {and} embedded devices.

\bibliographystyle{ACM-Reference-Format}
\bibliography{refs}

\appendix

\section{Evaluation on Unseen Test Set}
\label{sec:eval-unseen}
 
\begin{table}[!hbt]
\centering
\footnotesize
\setlength{\tabcolsep}{2.2pt}
\begin{tabular}{lrcccc}
\toprule
 & & & & \multicolumn{2}{c}{Five-shot} \\
\cmidrule(lr){5-6}
Task & Rows & Base & Tuned & Claude & DeepSeek \\
\midrule
BC5CDR          & 4,000 & 0.000 & \textbf{0.828} & 0.652 & 0.560 \\
Calendar        &   867 & 0.036 & \textbf{0.818} & 0.621 & 0.644 \\
TOPv2           & 5,000 & 0.002 & \textbf{0.562} & 0.221 & 0.112 \\
MultiCoNER      & 5,000 & 0.292 & \textbf{0.619} & 0.594 & 0.508 \\
DialogSum       &   459 & 0.241 & \textbf{0.451} & 0.387 & 0.347 \\
GoEmotions      & 4,000 & 0.393 & \textbf{0.525} & \textbf{0.515} & 0.466 \\
BEA-19          & 3,000 & 0.411 & \textbf{0.534} & 0.504 & \textbf{0.520} \\
CLINC (0.99)    & 4,000 & 0.704 & \textbf{0.882} & \textbf{0.894} & 0.833 \\
BFCL            & 1,234 & 0.808 & 0.816 & \textbf{0.870} & 0.848 \\
CLINC           & 4,000 & 0.111 & 0.822 & \textbf{0.894} & 0.833 \\
\bottomrule
\end{tabular}
\caption{\textnormal{Every model on the unseen test set, which the factory never used for training or for choosing a model. Base is the untrained model, and Tuned is the factory's fine-tuned model. Frontier models get the same five examples as in Table~\ref{tab:main}. Bold marks the best score and any score within the 0.015 noise floor of it. CLINC is intent classification; (0.99) marks the run with the goal pinned at 0.99. DialogSum's rows have one reference summary instead of three, so all its scores are lower than in Table~\ref{tab:main}.}}
\vskip -0.2in
\label{tab:unseen}
\end{table}

{We evaluated the models on unseen test sets that the factory never used for training or model selection: the fine-tuned model, its untrained base, and both frontier models given the same five examples as in Table~\ref{tab:main} (Table~\ref{tab:unseen}). } Spam detection's held-out set took the whole test split, leaving only five unused spam messages, so we could not check it on unseen rows.
 
{ The fine-tuned model beats its untrained base on every task, by 0.13 to 0.83, except function calling, where the untrained 4B model is already strong and fine-tuning adds 0.008. Most fine-tuned scores are lower than on the held-out set. On nested parsing, entity extraction and function calling, they drop by 0.035--0.037 while the untrained model barely moves. On intent classification and grammar correction, the untrained model drops just as much, so the unseen test set data are simply harder.}
 
{ Against frontier models given five examples, the factory's model wins five tasks on the unseen test set, by 0.024 to 0.342: biomedical extraction, calendar extraction, nested parsing, entity extraction and dialogue summarization. Three tasks are ties, within the noise floor: emotion classification, grammar correction, and intent classification with the goal at 0.99. Two are losses: function calling, and intent classification with the default goal, which stops on a 360M model. Compared with the held-out set, grammar correction moves from a win to a tie, and intent classification at goal 0.99 moves from a tie in our favor to one against.}


\section{Synthetic Data under Data Scarcity}
\label{sec:eval-scarce}
 
\begin{figure}[t]
  \centering
  \includegraphics[width=0.8\columnwidth]{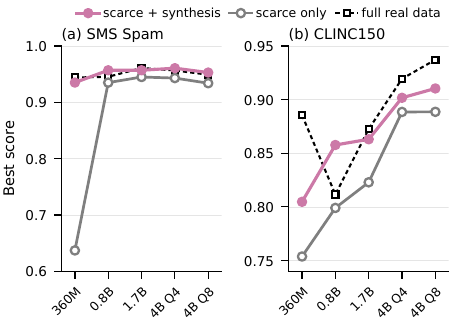}
 \vskip -0.15in
  \caption{\textnormal{Best score per tier, starting from (a) 100 and (b) 151 real rows, with and without synthesis, against full real data. The x-axis gives each tier's model.}}
  \vskip -0.15in
  \label{fig:scarce-pairs}
\end{figure}
 
    With thousands of real rows, synthetic rows made up only 3--22\% of each training set. So, we  starved two tasks of real data: 100 rows for spam detection, and 151 for intent classification, one per intent. Each task ran twice from the same rows, with mining off and the goal set to  0.99,  with and without synthesis data.



{ Synthesis won all ten tier comparisons} (Fig.~\ref{fig:scarce-pairs}), {eight of them by more than our 0.015 noise level and }four of them by more than 0.03. Tiers within a run share a growing training set, so they are not independent. How much synthesis helped depended on what limited the small model. 
 On spam detection, the 360M model scores 0.637 from 100 real rows and 0.935 with 362 generated rows added, close to its 0.944 on 4,045 {mostly} real rows. Without synthesis, only the 0.8B to 4B models reached that level. For the intent task, an  example for each of 151 intents is too few for any model. Synthesis added 0.013--0.059 at every tier and ended at 0.910, within 0.027 of the full-data run. Without it, no model passed 0.889.

We reran every task with only 100 real rows {(151 for intent classification, one per intent)}, {mining off} and synthesis enabled. {Nested parsing kept its full-data score, but its loader always adds a fixed 674-row in-domain adaptation set, so we leave it out of the counts that follow.} {Four of the other nine tasks} retained at least 97\% of their full-data score. Calendar extraction kept 98\%, but only on the 4B model; the 0.8B model fell 0.18 behind, suggesting real data matters more for smaller models. Grammar correction and summarization lost 7--10\%, while entity and emotion tasks lost 16--30\%. {Teacher accuracy did not predict this loss. The two largest drops came on the two tasks where the fewest synthetic rows passed quality control: 3\% for biomedical extraction and 24\% for fine-grained entities, against more than 60\% on every other task. Emotion classification points to a second limit: 99\% of its rows passed, but with 28 labels and 100 rows it starts with under four examples per label, and it lost 16\%.} This pass rate is cheap to estimate before relying on synthesis.

\section{Hardware requirements}
\label{app:settings}
 
Tables~\ref{tab:req-easy}--\ref{tab:req-hard} list the 60 requirements used in \S\ref{sec:eval-phone}. Each deploys one of our ten tasks in a way an app would use it, on one phone. A requirement sets four limits: peak memory, a minimum prefill speed, a minimum decode speed, and a maximum energy per request. A configuration meets the requirement only if it meets all four.
 
The kind of deployment sets the memory budget and the response time (Table~\ref{tab:req-kinds}). A fixed memory budget is capped at 25\% of the phone's RAM, which matters only on the 4\,GB Redmi 9A. The task's real request then sets the speeds: prefill must process the request's input within the first-token time, and decode must finish the answer within the remaining time. The fixed part of each prompt, such as instructions, label lists and tool schemas, is computed once and cached, so a request prefills only its input. A summary the user reads as it streams instead needs at least 5 tokens/s. The energy budget is a share of the phone's battery divided by the deployment's requests per day. The same deployment therefore has different limits on different phones.
 
The tier of a requirement is the share of the phone's configurations that meet it: easy if at least 60\% do, moderate if 30--60\% do, and hard if fewer than 30\% do. From the 38 deployments run on all four phones, we chose 20 requirements per tier, spread across tasks, phones and deployments, and kept only requirements that at least one configuration meets.
 
\begin{table}[h]
\centering
\small
\setlength{\tabcolsep}{4pt}
\begin{tabular}{lrrr}
\toprule
Deployment kind & Memory & First token & Answer \\
\midrule
Keyboard            & 300\,MB      & 0.3\,s & 1\,s \\
Voice assistant     & 1,536\,MB    & 0.5\,s & 2\,s \\
Feature in an app   & 1,024\,MB    & 0.5\,s & 3\,s \\
Background service  & 512\,MB      & 1\,s   & 5\,s \\
Dedicated app       & 40\% of RAM  & 1\,s   & 5\,s \\
Overnight job       & 50\% of RAM  & \multicolumn{2}{r}{6\,h $\div$ requests} \\
\bottomrule
\end{tabular}
\caption{Memory budget and response time by kind of deployment. An overnight job must finish all of a night's requests in 6 hours, at most 60\,s each.}
\label{tab:req-kinds}
\end{table}

\begin{table*}[t]
\centering
\small
\setlength{\tabcolsep}{4pt}
\begin{tabular}{llllrrrr}
\toprule
ID & Task & Phone & Deployment & Memory (MB) & Prefill & Decode & Energy (J) \\
\midrule
E01 & GoEmotions & S23 & Keyboard mood insights & $\le$ 3{,}500 & $\ge$ 1.9 & $\ge$ 1 & $\le$ 135 \\
E02 & SMS Spam & OP15 & Scam-check app & $\le$ 6{,}000 & $\ge$ 42 & $\ge$ 1 & $\le$ 101 \\
E03 & BEA-19 & OP8P & English journal corrections & $\le$ 5{,}800 & $\ge$ 0.8 & $\ge$ 1 & $\le$ 261 \\
E04 & DialogSum & S23 & Overnight call summaries & $\le$ 3{,}500 & $\ge$ 5.83 & $\ge$ 2 & $\le$ 1{,}350 \\
E05 & Calendar JSON & OP15 & Planner app & $\le$ 6{,}000 & $\ge$ 40 & $\ge$ 22 & $\le$ 101 \\
E06 & BFCL & OP15 & Shortcuts app & $\le$ 6{,}000 & $\ge$ 35 & $\ge$ 9 & $\le$ 50.6 \\
E07 & BC5CDR & OP15 & Paper reader app & $\le$ 6{,}000 & $\ge$ 32 & $\ge$ 12 & $\le$ 10.1 \\
E08 & MultiCoNER & OP15 & Photo and notes library index & $\le$ 7{,}600 & $\ge$ 4.07 & $\ge$ 7 & $\le$ 12.6 \\
E09 & TOPv2 & OP15 & Reminders and weather app & $\le$ 6{,}000 & $\ge$ 17 & $\ge$ 10 & $\le$ 50.6 \\
E10 & GoEmotions & OP8P & Keyboard mood insights & $\le$ 5{,}800 & $\ge$ 1.9 & $\ge$ 1 & $\le$ 157 \\
E11 & BEA-19 & S23 & English journal corrections & $\le$ 3{,}500 & $\ge$ 0.8 & $\ge$ 1 & $\le$ 225 \\
E12 & DialogSum & OP8P & Overnight call summaries & $\le$ 5{,}800 & $\ge$ 5.83 & $\ge$ 2 & $\le$ 1{,}566 \\
E13 & SMS Spam & OP15 & First-run inbox scan & $\le$ 7{,}600 & $\ge$ 19.4 & $\ge$ 1 & $\le$ 5.06 \\
E14 & BC5CDR & OP15 & Paper-alert abstract index & $\le$ 7{,}600 & $\ge$ 2.96 & $\ge$ 5 & $\le$ 25.3 \\
E15 & GoEmotions & OP15 & Journaling app mood tags & $\le$ 6{,}000 & $\ge$ 57 & $\ge$ 2 & $\le$ 25.3 \\
E16 & BEA-19 & OP15 & English-learning feedback & $\le$ 6{,}000 & $\ge$ 24 & $\ge$ 5 & $\le$ 20.2 \\
E17 & DialogSum & OP15 & Meeting-notes app & $\le$ 6{,}000 & $\ge$ 175 & $\ge$ 5 & $\le$ 101 \\
E18 & BEA-19 & OP15 & Essay proofreading app & $\le$ 6{,}000 & $\ge$ 24 & $\ge$ 5 & $\le$ 33.7 \\
E19 & GoEmotions & OP15 & Keyboard mood insights & $\le$ 7{,}600 & $\ge$ 1.9 & $\ge$ 1 & $\le$ 253 \\
E20 & DialogSum & OP15 & Overnight call summaries & $\le$ 7{,}600 & $\ge$ 5.83 & $\ge$ 2 & $\le$ 2{,}529 \\
\bottomrule
\end{tabular}
\caption{Easy requirements: at least 60\% of the phone's configurations meet all four limits. Prefill and decode are minimum speeds in tokens/s. OP15 = OnePlus 15, S23 = Galaxy S23, OP8P = OnePlus 8 Pro, R9A = Redmi 9A.}
\label{tab:req-easy}
\end{table*}
 
\begin{table*}[t]
\centering
\small
\setlength{\tabcolsep}{4pt}
\begin{tabular}{llllrrrr}
\toprule
ID & Task & Phone & Deployment & Memory (MB) & Prefill & Decode & Energy (J) \\
\midrule
M01 & CLINC150 & OP15 & Offline assistant app & $\le$ 6{,}000 & $\ge$ 515 & $\ge$ 1 & $\le$ 33.7 \\
M02 & GoEmotions & OP8P & Comment tone check & $\le$ 1{,}024 & $\ge$ 114 & $\ge$ 2 & $\le$ 31.3 \\
M03 & SMS Spam & S23 & Scam-check app & $\le$ 2{,}800 & $\ge$ 42 & $\ge$ 1 & $\le$ 54 \\
M04 & BEA-19 & R9A & English journal corrections & $\le$ 1{,}900 & $\ge$ 0.8 & $\ge$ 1 & $\le$ 290 \\
M05 & BFCL & OP15 & Voice-assistant actions & $\le$ 1{,}536 & $\ge$ 70 & $\ge$ 24 & $\le$ 50.6 \\
M06 & Calendar JSON & S23 & Planner app & $\le$ 2{,}800 & $\ge$ 40 & $\ge$ 22 & $\le$ 54 \\
M07 & BC5CDR & OP8P & Paper-alert abstract index & $\le$ 5{,}800 & $\ge$ 2.96 & $\ge$ 5 & $\le$ 15.7 \\
M08 & DialogSum & R9A & Overnight call summaries & $\le$ 1{,}900 & $\ge$ 5.83 & $\ge$ 2 & $\le$ 1{,}737 \\
M09 & MultiCoNER & OP15 & Search query understanding & $\le$ 1{,}024 & $\ge$ 44 & $\ge$ 14 & $\le$ 10.1 \\
M10 & TOPv2 & S23 & Reminders and weather app & $\le$ 2{,}800 & $\ge$ 17 & $\ge$ 10 & $\le$ 27 \\
M11 & BEA-19 & OP8P & Email ``fix grammar'' button & $\le$ 1{,}024 & $\ge$ 48 & $\ge$ 8 & $\le$ 31.3 \\
M12 & GoEmotions & R9A & Keyboard mood insights & $\le$ 1{,}900 & $\ge$ 1.9 & $\ge$ 1 & $\le$ 174 \\
M13 & BFCL & OP15 & Smart-home app commands & $\le$ 1{,}024 & $\ge$ 70 & $\ge$ 14 & $\le$ 25.3 \\
M14 & SMS Spam & S23 & First-run inbox scan & $\le$ 3{,}500 & $\ge$ 19.4 & $\ge$ 1 & $\le$ 2.7 \\
M15 & MultiCoNER & OP8P & Photo and notes library index & $\le$ 5{,}800 & $\ge$ 4.07 & $\ge$ 7 & $\le$ 7.83 \\
M16 & BC5CDR & OP15 & Reference-manager highlight tags & $\le$ 1{,}024 & $\ge$ 64 & $\ge$ 18 & $\le$ 25.3 \\
M17 & DialogSum & S23 & Meeting-notes app & $\le$ 2{,}800 & $\ge$ 175 & $\ge$ 5 & $\le$ 54 \\
M18 & TOPv2 & OP8P & Reminders and weather app & $\le$ 4{,}600 & $\ge$ 17 & $\ge$ 10 & $\le$ 31.3 \\
M19 & MultiCoNER & OP15 & Read-later entity tags & $\le$ 6{,}000 & $\ge$ 22 & $\ge$ 9 & $\le$ 5.06 \\
M20 & BFCL & S23 & Shortcuts app & $\le$ 2{,}800 & $\ge$ 35 & $\ge$ 9 & $\le$ 27 \\
\bottomrule
\end{tabular}
\caption{Moderate requirements: 30--60\% of the phone's configurations meet all four limits. Prefill and decode are minimum speeds in tokens/s.}
\label{tab:req-moderate}
\end{table*}
 
\begin{table*}[t]
\centering
\small
\setlength{\tabcolsep}{4pt}
\begin{tabular}{llllrrrr}
\toprule
ID & Task & Phone & Deployment & Memory (MB) & Prefill & Decode & Energy (J) \\
\midrule
H01 & CLINC150 & S23 & Voice-assistant intent router & $\le$ 1{,}536 & $\ge$ 1{,}030 & $\ge$ 3 & $\le$ 27 \\
H02 & SMS Spam & OP15 & Messaging app spam check & $\le$ 512 & $\ge$ 42 & $\ge$ 1 & $\le$ 10.1 \\
H03 & Calendar JSON & OP8P & Voice ``add to my calendar'' & $\le$ 1{,}536 & $\ge$ 80 & $\ge$ 57 & $\le$ 104 \\
H04 & BFCL & R9A & Smartwatch actions relay & $\le$ 512 & $\ge$ 35 & $\ge$ 9 & $\le$ 17.4 \\
H05 & GoEmotions & S23 & Keyboard tone check & $\le$ 300 & $\ge$ 190 & $\ge$ 8 & $\le$ 5.4 \\
H06 & BEA-19 & OP15 & English-learner keyboard & $\le$ 300 & $\ge$ 80 & $\ge$ 29 & $\le$ 5.06 \\
H07 & MultiCoNER & OP8P & Long-press smart actions & $\le$ 512 & $\ge$ 22 & $\ge$ 9 & $\le$ 15.7 \\
H08 & TOPv2 & R9A & Reminders app quick add & $\le$ 950 & $\ge$ 34 & $\ge$ 16 & $\le$ 34.7 \\
H09 & DialogSum & S23 & Dialer call summary & $\le$ 1{,}024 & $\ge$ 350 & $\ge$ 5 & $\le$ 90 \\
H10 & BC5CDR & OP8P & Paper reader app & $\le$ 4{,}600 & $\ge$ 32 & $\ge$ 12 & $\le$ 6.26 \\
H11 & CLINC150 & OP15 & Banking app assistant & $\le$ 1{,}024 & $\ge$ 1{,}030 & $\ge$ 2 & $\le$ 506 \\
H12 & TOPv2 & R9A & Smartwatch command relay & $\le$ 512 & $\ge$ 17 & $\ge$ 10 & $\le$ 23.2 \\
H13 & Calendar JSON & S23 & Calendar quick add & $\le$ 1{,}024 & $\ge$ 80 & $\ge$ 34 & $\le$ 54 \\
H14 & DialogSum & OP15 & Chat catch-up & $\le$ 1{,}024 & $\ge$ 350 & $\ge$ 5 & $\le$ 50.6 \\
H15 & SMS Spam & OP8P & Messaging app spam check & $\le$ 512 & $\ge$ 42 & $\ge$ 1 & $\le$ 6.26 \\
H16 & BEA-19 & R9A & English-learning feedback & $\le$ 1{,}500 & $\ge$ 24 & $\ge$ 5 & $\le$ 13.9 \\
H17 & BFCL & S23 & Voice-assistant actions & $\le$ 1{,}536 & $\ge$ 70 & $\ge$ 24 & $\le$ 27 \\
H18 & GoEmotions & OP15 & Keyboard tone check & $\le$ 300 & $\ge$ 190 & $\ge$ 8 & $\le$ 10.1 \\
H19 & BC5CDR & OP8P & Reference-manager highlight tags & $\le$ 1{,}024 & $\ge$ 64 & $\ge$ 18 & $\le$ 15.7 \\
H20 & MultiCoNER & R9A & Long-press smart actions & $\le$ 512 & $\ge$ 22 & $\ge$ 9 & $\le$ 17.4 \\
\bottomrule
\end{tabular}
\caption{Hard requirements: under 30\% of the phone's configurations meet all four limits. Prefill and decode are minimum speeds in tokens/s.}
\label{tab:req-hard}
\end{table*}

\end{document}